# 六足机器人动态目标检测与跟踪系统研究

王德鑫

（山东大学控制科学与工程学院， 山东 济南 250061）

**摘 要**：动态目标检测与目标跟踪是图像领域的热点研究问题，为探索其在移动机器人领域的应用价值，本文基于六足机器人，设计了动态目标检测与跟踪系统。首先基于运动背景补偿法，提出区域合并与自适应外点滤除相结合的动态目标检测方法，该方法通过对称匹配、自适应外点滤除实现了运动背景的精确补偿，并通过区域合并实现非刚体目标的完整检测。其次研究了基于 KCF 的目标跟踪算法在六足机器人平台上的应用，通过自适应调整跟踪速度实现六足机器人对运动目标的角度跟踪。最后设计了由操作人员、处理器、六足机器人和视觉传感器等组成的系统架构，并将本文提出的运动目标检测及跟踪算法应用于系统中。实验表明，将改进后的算法应用于移动六足机器人的系统中，对运动目标可实现有效的检测与跟踪。

**关键词**：六足机器人；运动补偿；动态目标检测；目标跟踪

**中图分类号**：TP242.6    **文献标识码**：A

## Research on dynamic target detection and tracking system of hexapod robot

WANG Dexin

（*The School of Control Science and Engineering (CSE) of Shandong University, Jinan 250061, China*）

**Abstract:** Dynamic target detection and target tracking are hot issues in the field of image. In order to explore its application value in the field of mobile robot, a dynamic target detection and tracking system is designed based on hexapod robot. Firstly, the dynamic target detection method is introduced with region merging and adaptive external point filtering based on motion compensation method. This method achieves the accurate compensation of the moving background through symmetric matching and adaptive external point filtering, and achieves complete detection of non-rigid objects by region merging. Secondly, the application of target tracking algorithm based on KCF in hexapod robot platform is studied, and the Angle tracking of moving target is realized by adaptive adjustment of tracking speed. The last, the architecture of robot monitoring system is designed, which consists of operator, processor, hexapod robot and vision sensor, and the moving object detection and tracking algorithm proposed in this paper is applied to the system. The experimental results show that the improved algorithm can effectively detect and track the moving target when applied to the system of the mobile hexapod robot.

## 1 引言（Introduction）

基于机器人平台的动态目标检测与跟踪系统可用于多种场景[1-2]，六足机器人以其适应复杂地形的特点[3]，可在复杂环境中执行多种任务，具有较大的研究价值。

自 1989 年，麻省理工学院研制出第一个真正意义上的六足机器人 Genghis 开始，六足机器人开始进入大众视野[4-5]。六足机器人的着地点分散，使机器人在凹凸不平的地面上如履平地，并有效降低机器人在运动中的噪声和机器人本体的晃动，使六足机器人在执行监视、勘测、跟踪等任务时不易被发现。

很多场合下需要机器人在运动过程中检测动态目标并跟踪，如林区防盗伐、厂区防盗、移动监控[6]等。动态目标检测目前主要有运动补偿法[7]和光流法[8]。运动补偿法具有减少图像畸变等优点，但目前没有一个较好的算法可以准确计算背景的运动参数，且结果有较大的噪声。光流法的优点是不需要对图像进行预处理和能处理运动目标重叠、遮挡等问题，缺点是计算量大。LUO Xuhao 等[9]通过小波多分辨率分析和相邻帧间特征点位置估计改进 SIFT 方法，实现快速全局背景运动补偿参数估计，提高参数估计精确性，加快了特征配准和检测速度，但是对运动缓慢的目标检测效果较差。Qin Yue[10] 等结合 LK，局部方差，局部熵和聚类方法实现准确得检测到动态目标并降低了时间复杂度，但是对于小目标的检测效果较差。

本文在结合前人工作的基础上，改进了运动目标检测算法，通过对称匹配、自适应外点滤除[11]、区域合并等方法提高了检测的完整性，并基于六足机器人设计了自适应调整运动速度的跟踪算法，将

其应用在六足机器人上，结合无线通信、人机交互等技术，开发了一套具有安全可靠的感知能力、执行能力的系统。六足机器人的步态规划技术已趋于成熟，不在本文研究范围内。

## 2 动态目标检测与跟踪算法研究

### 2.1 动态目标检测

由于六足机器人在运动过程中无法精确保持机体的平稳，所以搭载在机器人上的视觉传感器拍摄的图像会出现抖动的现象。为保证机器人在图像抖动情况下对运动目标的精确检测，本文采用对称匹配、自适应外点滤除和区域合并相结合的运动目标检测算法。

首先使用 SURF 特征[12]在第 $n$ 帧和第 $n-1$ 帧图像中检测特征点 $Ps_n$ 和 $Ps_{n-1}$，分别对 $Ps_n$ 和 $Ps_{n-1}$ 计算在另一帧中匹配到的特征点 $Ps'_{n-1}$ 和 $Ps'_n$，将在两次匹配过程中都匹配到的特征点对记为正确的匹配点对 $Ps$，上述过程称之为对称匹配。匹配方法采用 KNN 匹配[13]。

特征点对 $Ps$ 包含静止背景点和运动的前景点，运动的前景点对在运动补偿时会使结果产生较大误差，所以要滤除前景点对，只保留背景点对进行运动补偿。使用自适应外点滤除[11]将特征点中的背景点与前景点分离，进而只保留背景点进行仿射变换，提高运动补偿的精度。

使用高斯滤波将第 $n-1$ 帧与运动补偿后的第 $n$ 帧图像去除噪声，再通过帧差法粗略去除背景，将结果进行二值化处理，前景像素设为 255，背景像素设为 0。二值图形会存在一些噪点，通过膨胀腐蚀去除部分离群点和较小的噪点。对于分辨率为 640*480 的视频序列，形态学操作模板大小设置为 3*3。

由于运动目标多为非刚体，目标不同部分的运动存在差异性，导致一个目标被检测为多个分散区域，且当多个目标同步运动时无法区分。本文提出的区域合并方法可提高非刚体目标的检测完整性。

(1) 当前帧内的归并。若同一帧内的两个区域质心小于阈值 $th1$ 且 HSV 颜色空间的欧氏距离小于阈值 $th2$，则将这两个区域设置为连通域。

(2) 连续帧内的归并。若相邻帧内两个目标的质心距离小于阈值 $th3$ 且 HSV 空间欧式距离小于阈值 $th4$，设这两个区域为等价区域对。迭代后可在连续两帧内获得多组等价区域对 $\{(S_n^i, S_{n-1}^i), i \leq I\}$，I 为等价区域对数。计算每一个等价区域对的运动矢量，然后计算任意两对等价区域对的运动一致性 (Motion Consistency, MC)，若两区域对的欧氏距离小于阈值 $th5$ 且运动一致性小于阈值 $th6$，则两对等价区域对中在当前帧的区域设置为连通域。

$$MC = \sqrt{(dx_n - dx_{n-1})^2 + (dy_n - dy_{n-1})^2} \quad (1)$$

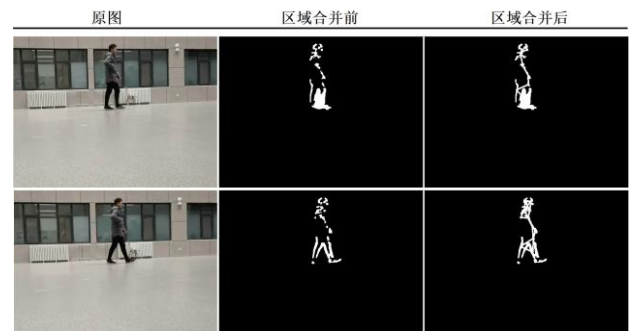

图 1 区域合并

区域合并算法内的阈值全部由 VTB100 数据集中的视频测试得到(th1=30，th2=3000，th3=30，th4=8000，th5=50，th6=30)。

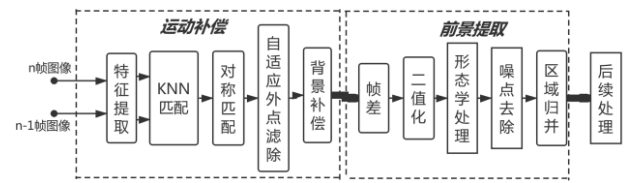

图 2 动态目标检测流程图

### 2.2 基于六足机器人的目标跟踪

六足机器人与轮式机器人的不同之处在于，轮式机器人只需简单的算法控制电机即可转动到任意角度，而六足机器人一次转向，则需要至少一个步态周期，即腿足的抬起、转动和落下，耗时较长且控制更加复杂。本文在三角步态的基础上，设计了自适应调整转动速度的跟踪方法，实现六足机器人对运动目标的快速跟踪，始终保持目标在视野中央附近。六足机器人检测到运动目标后，由操作人员选定目标进行跟踪。

表 1 对比了近几年目标跟踪的经典算法，测试对象为 VTB-100 数据集中的部分视频。视频 1(V1) 中背景简单，目标较大且特征明显；视频 2(V2)背景简单，目标较小且与背景颜色接近；视频 3(V3)

背景与目标颜色相近，有多个目标，目标较小且存在遮挡；视频 4(V4)背景简单，目标特征明显，但形状和尺度变化较大；视频 5(V5)背景复杂，目标较多且存在目标遮挡。√表示在跟踪过程中无跟丢现象，×表示有跟丢现象。结果显示，KCF 算法[15]对形状和尺度变化具有较强的鲁棒性，且帧率满足实时性要求,满足六足机器人跟踪运动目标的需求，所以本文基于 KCF 跟踪算法进行设计。

表 1 跟踪算法比较
Tab.1 Tracking algorithm comparison

| 算法 | V1 | V2 | V3 | V4 | V5 | 帧数/s |
|---|---|---|---|---|---|---|
| CAMShift | √ | × | × | × | × | 250 |
| KCF | √ | √ | × | √ | × | 53 |
| Boost | √ | √ | × | √ | × | 30 |
| MIL | √ | × | × | √ | × | 24 |
| MedianFlow | √ | × | × | √ | × | 28 |
| TLD | √ | √ | × | × | × | 25 |

综合考虑运动速度和稳定性，本文采用占地系数为 0.5 的三角步态[16]，即每条腿的接触地面时间占步态周期的 0.5 倍。将机器人的六个腿部结构分为 R 和 L 两组，直行步态可分为 R 抬升、L 前移、R 落下、L 抬升、R 前移、L 落下 6 个步骤，循环执行。首先将上述 6 组动作在实验环境中分别编程实现并保存为函数，按顺序组合后运行可实现六足机器人的直行。转弯步态包括沿中心转弯和沿外点转弯。本文采用沿中心转弯，即要求机器人直行停止后再转弯，过程与直行步态类似，不再赘述。

六足机器人一次转弯的弧度 $d\theta$ 与 $x$ 轴偏移量 $dx$ 的关系如式(2)所示。$dx$ 单位为 pt（像素）。

$$d\theta = \begin{cases} \dfrac{dx}{|dx|} \cdot (|dx| - th) \cdot k, & |dx| > th \\ 0, & |dx| < th \end{cases} \quad (2)$$

其中， $th = 640/(4*2) = 80 pt$ ，
$k = \dfrac{\pi/12}{320-80} = 2.18e-3$。

当目标在图像中存在 $y$ 轴方向的偏差 $dy$ 时，摄像头云台的抬高与下降角度计算方法与式(2)类似，只需将 320 改为 240 即可。图 3 为跟踪偏差示意图，图中 $(dx, dy)$ 的数据范围分别是 $(-320 \sim 320 pt, -240 \sim 240 pt)$。

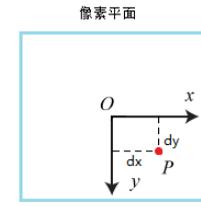

图 3 跟踪偏移示意图

## 3 基于六足机器人的动态目标检测与跟踪系统设计

### 3.1 系统体系结构

六足机器人需要完成的任务包括运动控制、动态目标检测、目标跟踪、数据传输。除此之外，需要有操作人员远程操控，防止机器人决策错误或无法决策，保证系统的稳定、可靠。系统结构如图 4 所示。

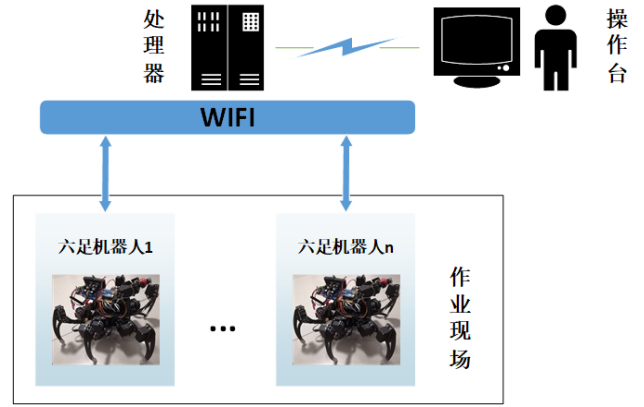

图 4 系统结构

操作人员通过显示器实时查看机器人运动状态、检测与跟踪情况，对机器人有最高优先级的决策，可通过键盘手动控制机器人的行进方向和摄像头的角度，并可中断机器人的跟踪过程。

处理器负责实时接受图像数据、检测动态目标、跟踪目标以及下达运动指令,协调整个系统的运行。

六足机器人由机器人主体、微型控制器、通信系统、图像系统和电源组成，负责向处理器发送实时拍摄的图像数据，并接受运动控制指令，将其解析之后驱动腿关节舵机转动，带动机器人移动。

### 3.2 系统架构和硬件组成

系统架构主要包括操作系统层、驱动层和硬件层三个部分。

硬件层由六足机器人主体、相机、电机和电源组成。主体采用菱形而不是长方形，以增加腿部结构的灵活性和稳定性，为兼顾轻便和结实，机器人的主体和 6 个腿部结构均采用硬铝合金材料，驱动关节采用 20 个 MG995 舵机，包括 18 个腿部关节和 2 个摄像头云台关节。相机为 FOXEER 公司的 Monster micro Pro 摄像头，采用五百万像素无畸变镜头。电源采用 12V 的格式锂电池，

向 20 个舵机和其他模块供电，工作电流最高可达 6A，ESP8266 通信单元由单片机的 3.3V 引脚直接供电。

驱动层作为连通操作系统层与硬件层的中间环节，主要完成传感器信息的传输以及控制指令的转达。图像预处理模块首先接收相机拍摄的初始图像，进行简单预处理后发送到操作系统的图像数据采集模块，经过一系列处理后通过 wifi 无线通信单元 ESP8266 转达至机器人的处理器，经处理器解算控制指令后，将解算后的电机转速及转角指令发送至电机控制器，由电机控制器直接控制电机的运动。考虑到本文不涉及运动控制的研究，不需要在机器人端进行复杂的运动控制解析，机器人处理器采用 STM32F103RCT6 单片机。采用 TS832+RS832 图像传输模块，图像分辨率为 640*480。六足机器人的驱动装置共包括腿关节和云台关节的 20 个舵机，采用 24 路舵机控制板作为驱动控制器，单片机只需向舵机控制板发送连续的字符串数据，即可同时控制 20 个关节。

操作系统层是控制核心，集成了机器人的功能模块，负责采集图像、图像处理、人机交互等。为与图像传输接口保持一致，本文选用 Windows10 操作系统，运行于图 4 中的处理器中

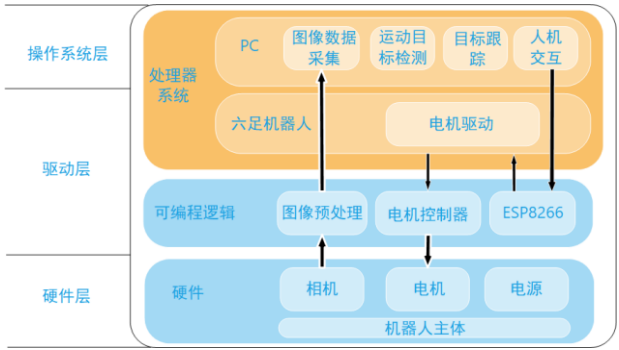

图 5 系统总体架构

### 3.3 软件框架

软件结构采用模块化设计方式，如图 7 所示。处理器内的软件系统由顶层的人机交互和底层的智能算法组成。人机交互模块负责显示各项数据、接收手动控制指令（由键盘实时操控）等任务，智能算法模块负责各项图像处理算法、保存数据以及将控制指令转码为本文自定义的 RCP（Robot communication protocol）通信协议，RCP 通信协议如图 6 所示。各模块之间使用多线程的方式运行，避免程序的阻塞，保证了处理器实时、稳定的协调所有模块的运行。

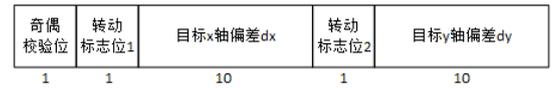

图 6 RCP 通信协议

RCP 通信协议由 21 位二进制数组成。转动标志位 1 和 2 分别表示是否转弯和是否转动摄像头云台，标志位为 1 时有效。$(dx, dy)$ 的含义请看 3.4 节，每个变量用 10 位二进制数表示，其中第高位为符号位，置 1 表示右转，置 0 表示左转，可表示的偏移范围为 $(-512pt, 512pt)$。

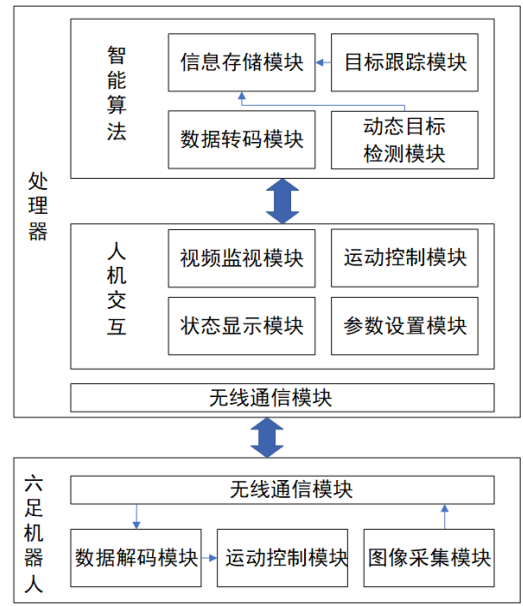

图 7 系统软件结构

六足机器人内的软件系统由数据解码模块、运动控制模块和图像采集模块组成。数据解码模块将处理器发送的 RCP 数据解码并计算出运动控制信息，运动控制模块则直接控制电机转动。无线通信模块由基于 ESP8266 的 WIFI 通信和 5.8G 图像传输模块组成，两种通信的功能请回顾 3.2 节。六足机器人内的每个模块在独立的微型处理器上运行，只需对接收的数据处理后发出，无需考虑其他模块的运行情况，互不干扰，保证了机器人各模块的稳定运行。

### 3.4 人机交互平台

人机交互平台由视频监视模块、运动控制模块、状态显示模块、无线通信模块和参数设置模块组成。操作人员可实时查看机器人的运行状态、无线通信状态、切换图像模式以及手动控制机器人的运动。人机交互界面如图 8。

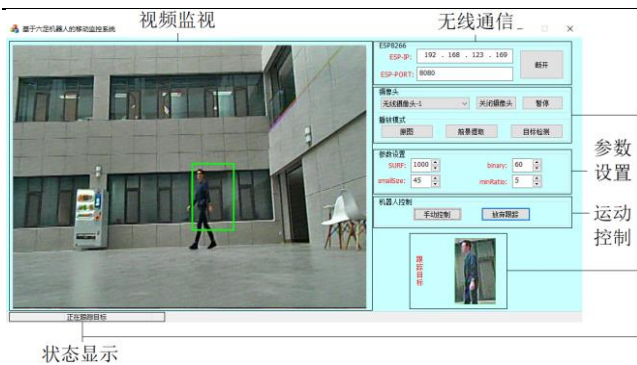

图 8 人机交互界面

如果检测与跟踪算法的参数对当前场景不适用，操作人员可通过设置算法参数优化算法，提高检测与跟踪的精度。

### 3.5 实验环境

实验环境如下所示：

硬件：intel skull nuc，core i7-6770HQ，2.6-3.5 GHz.

操作系统：windows 10.

编程环境：Visual Studio 2017，Keil v5，C++，opencv3.4.0, opencv-contrib3.4.0.

## 4 实验过程（The experimental process）

为测试六足机器人在移动过程中对运动目标的检测性能，在实验楼大厅进行试验，背景复杂度适中，六足机器人运动速度和目标运动速度都在每帧 10 个像素以内。首先在人机交互界面手动控制六足机器人在大厅内随机移动，由两名同学模拟运动目标由视野外进入视野并相对行走，令由一名同学静坐不动。观察六足机器人是否能在运动过程中检测到两位运动的同学。检测结果如图 10 所示，检测帧率如表 2。

由检测结果可以看到，两位运动的同学均被完整地检测到，并用绿色矩形框标记出来，检测帧率在复杂场景下可达到 17.5 帧/s。但其中存在将背景误检测的情况，我们用动态目标检测程序来检测 VTB100 数据集，检测结果较机器人检测的误检率小。我们分析试验流程及硬件设备后认为，六足机器人没有集成运动控制算法以及采用的舵机存在"抖舵"的现象，使得机器人在运动过程中会有晃动及轻微的抖动，导致相机拍摄的图片出现大面积模糊，所以实际运行过程中的误检率较高。

为测试移动监控系统对动态目标的跟踪性能，由一名同学模拟运动目标，在相机视野内运动，首先手动在人机交互界面选中待跟踪目标，点击"自动跟踪"按键后，六足机器人开始自动跟踪目标。图 9 中绘制了连续 330 帧跟踪过程中的 x 轴和 y 轴方向的像素偏移，目标在第 0 帧出现在视野边缘，在第 120 帧向视野边缘运动，在第 250 帧开始反向运动。从图 9 中可以看出，当目标与视野中央的偏差大于 80pt 时，机器人快速转动，使目标回到视野中央附近，跟踪延迟约为 1s。图 11 为跟踪结果，图 11(a)最右图为手动选定的跟踪目标。

由表 2 可得，跟踪帧率在不同复杂程度的场景下并没有较大差距，而检测帧率相差 6.8 帧，这是由于复杂场景中的特征点比简单场景多，检测算法需要检测并匹配特征点来进行运动补偿造成的。

表 2 检测与跟踪帧率（帧/秒）
Tab.2 Detection and tracking of frame rate (frame/s)

| 场景 | 检测 | 跟踪 |
| --- | --- | --- |
| 复杂场景 | 17.5 | 42.1 |
| 简单场景 | 24.3 | 45.3 |

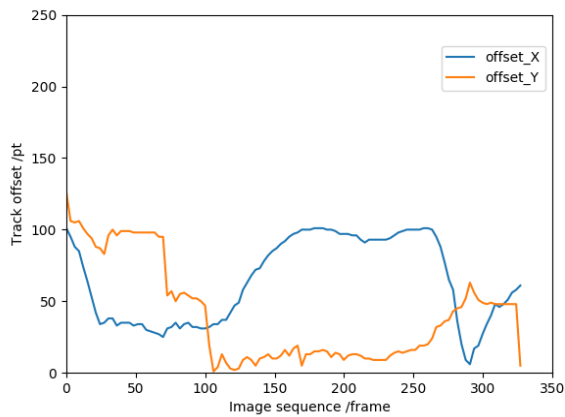

图 9 目标跟踪偏移

## 5 结论（Conclusion）

为探索动态目标检测与跟踪算法与机器人的结合应用，本文基于六足机器人平台，结合人机交互、智能算法、智能机器人等于一体开发了一套完整的检测与跟踪系统。其中，针对以往动态目标检测算法易将目标检测为多个分散区域的问题，通过区域合并与运动补偿法相结合，使其在六足机器人在运动过程中可以完整检测非刚体运动目标。采用基于 KCF 的目标跟踪算法,通过自适应调整跟踪速度，实现六足机器人稳定可靠地跟踪运动目标。

本文后续将在此基础上研究六足机器人三角步态规划与动态目标跟踪相融合的运动控制算法，以提高跟踪精度和机器人在运动过程中的平稳性。

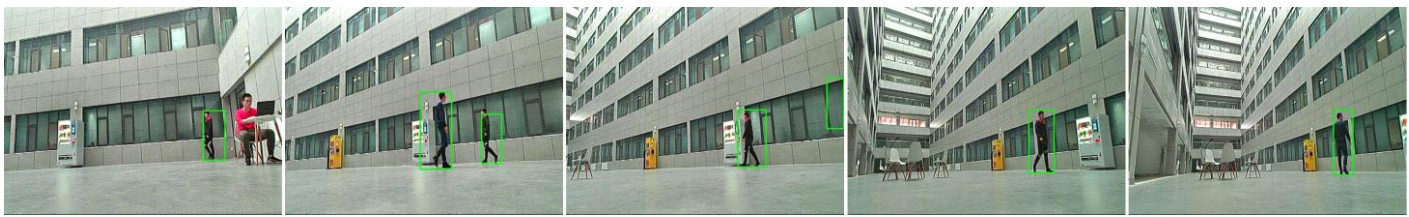

图 10 动态目标检测结果

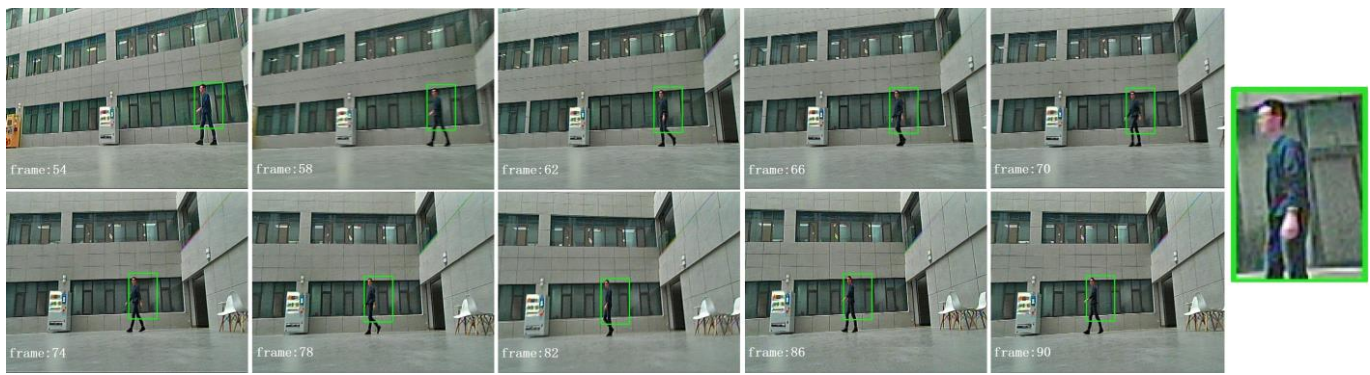

(a)

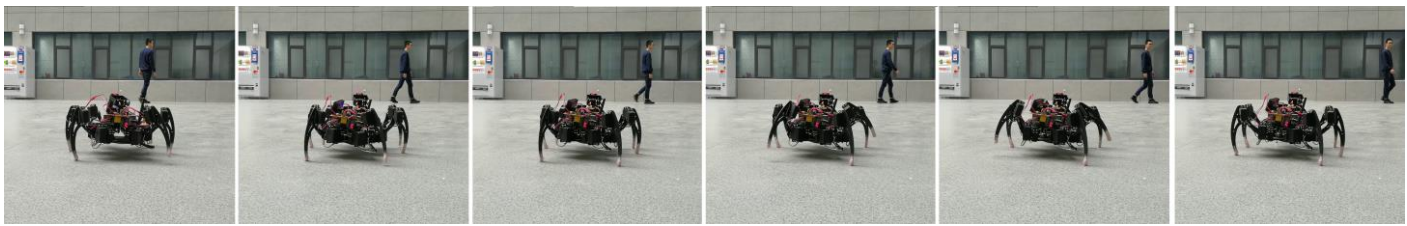

(b)

图 11 目标跟踪